\documentclass[conference]{IEEEtran}
\IEEEoverridecommandlockouts
\usepackage[top=0.75in, bottom=1.05in, left=0.635in, right=0.635in]{geometry}
\usepackage{cite}
\usepackage{amsmath,amsfonts}
\usepackage{bbm}
\usepackage{graphicx}
\usepackage{textcomp}
\usepackage{lipsum}
\usepackage{xcolor}
\usepackage{mathtools}
\usepackage{float}
\usepackage{algorithm}
\usepackage{algpseudocode}
\usepackage{amssymb}
\usepackage{caption}
\usepackage{array}
\newcolumntype{d}[1]{>{\centering\arraybackslash}m{#1\linewidth}}

\DeclareMathOperator*{\argmax}{argmax}
\def\BibTeX{{\rm B\kern-.05em{\sc i\kern-.025em b}\kern-.08em
T\kern-.1667em\lower.7ex\hbox{E}\kern-.125emX}}
\begin{document}

    \title{\huge Graph Reinforcement Learning for QoS-Aware\\ Load Balancing in Open Radio Access  Networks
\thanks{© 2025 IEEE. Personal use of this material is permitted. Permission from IEEE must be obtained for all other uses, in any current or future media, including reprinting/republishing this material for advertising or promotional purposes, creating new collective works, for resale or redistribution to servers or lists, or reuse of any copyrighted component of this work in other works.}
  }

    \author{\IEEEauthorblockN{
        Omid Semiari
        , Hosein Nikopour
        and Shilpa Talwar
    }\IEEEauthorblockA{
        Intel Labs, Santa Clara, CA\\Email: \{omid.semiari, hosein.nikopour, shilpa.talwar\}@intel.com\vspace{-1em}}
    }

    \maketitle

    \begin{abstract}
        Next-generation wireless cellular networks are expected to provide unparalleled Quality-of-Service (QoS) for emerging wireless applications, necessitating strict performance guarantees, e.g., in terms of link-level data rates. A critical challenge in meeting these QoS requirements is the prevention of cell congestion, which involves balancing the load to ensure sufficient radio resources are available for each cell to serve its designated User Equipments (UEs).  
        In this work, a novel \emph{QoS-aware Load Balancing (LB)} approach is developed to optimize the performance of Guaranteed Bit Rate (GBR) and Best Effort (BE) traffic in a multi-band Open Radio Access Network (O-RAN) under QoS and resource constraints. 
        The proposed solution builds on Graph Reinforcement Learning (GRL), a powerful framework at the intersection of Graph Neural Network (GNN) and RL. The QoS-aware LB is modeled as a Markov Decision Process, with states represented as graphs. QoS consideration are integrated into both state representations and reward signal design. The LB agent is then trained using an off-policy dueling Deep Q Network (DQN) that leverages a GNN-based architecture.  This design ensures the LB policy is invariant to the ordering of nodes (UE or cell), flexible in handling various network sizes, and capable of accounting for spatial node dependencies in LB decisions. Performance of the  GRL-based solution is compared with two baseline methods. Results show substantial performance gains,  including a $53\%$ reduction in QoS violations and a fourfold increase in  the 5th percentile rate  for  BE traffic.\vspace{-.5em}
    \end{abstract}


    \section{Introduction}~\label{sec:intro}\vspace{-1.0em}
   
    Next-generation cellular networks are anticipated to provide a reliable communications and compute backbone for enabling  emerging wireless applications such as unbounded Extended Reality (XR), online video streaming with extreme $8$K resolution, and a suite of services driven by generative image and language models (e.g., conversational AI)\cite{8869705}. A  common factor  among these services is the need for a Guaranteed Bit Rate (GBR) between access points and User Equipments (UEs). GBR is a key Quality-of-Service (QoS) class that significantly influences users' perception of service quality. 
    
    Supporting GBR entails provisioning resources and managing traffic loads at each cell to prevent congestion, ensuring that each cell can deliver GBR traffic to its assigned UEs. In overloaded networks, \emph{load balancing (LB)} is a promising technique to offload UEs from congested cells to nearby under-loaded cells. By carefully designing LB constraints, such forced handovers can free up resources to alleviate cell congestion without compromising the performance of offloaded UEs. However, LB is not trivial when dealing with QoS traffic. QoS-aware LB is particularly challenging in real-world scenarios due to the mixed GBR requirements from different UEs, along with non-deterministic factors such as fading channel, interference, UE mobility, and traffic variations.
    
    A large body of work in~\cite{8255801, 10659897, DEMIRCI2018140, 10723743, 10615828, 9740126, 10463334, 8796404, 9369525, 9637122, 9679991} has focused on addressing LB in radio access network (RAN). Some of the model-based techniques for LB include cell range expansion via adjusting handover parameters such as Cell Individual Offset (CIO) values~\cite{8255801, 10659897}, cell breathing techniques~\cite{DEMIRCI2018140}, and  threshold-based traffic steering~\cite{10723743}. While interesting, LB policies in~\cite{8255801, 10659897, DEMIRCI2018140, 10723743} are designed based on long-term traffic patterns and too simple to deal with UE-level QoS constraints in  real-world networks. The work in~\cite{10615828} develops an LB policy via linear programming, however the scope is limited to intra-site LB. Moreover, the non-deterministic policy in~\cite{10615828} cannot provide rate guarantees due to its probabilistic nature.  
    
    Load balancing is recognized as one of the key use cases of Machine Learning (ML) for RAN optimization~\cite{lin2023}. The authors in~\cite{9740126} provide a  summary of ML-based methods for LB.~\cite{10463334} introduces a recurrent neural network to perform LB based on UEs' mobility predictions. The authors in~\cite{8796404} adopt a cell clustering scheme based on historic data and subsequently apply a Deep Reinforcement Learning (DRL) method for LB within each cluster. The works in~\cite{9369525} and \cite{9637122} develop different DRL-based policies to optimize CIO for LB. Despite presenting novel ideas, most of prior works in~\cite{9740126, 10463334, 8796404, 9369525, 9637122} focus on cell-level information and do not account for integrating per-UE QoS constraints into their design. In addition, the deep neural networks proposed in these works face scalability issues, as they take fixed-sized inputs and therefore do not generalize to RAN environments with different topologies.

    Recently, Graph Reinforcement Learning (GRL) has been adopted to solve various networking problems~\cite{zhao2023}, including LB~\cite{9679991}. In GRL, Graph Neural Network (GNN) can serve as the underlying architecture for the DRL agent's policy or value networks, motivated by the fact that RAN states can be modeled as graphs. Compared to other architectures, GNN offers unique capabilities, including: (a) flexibility to scale to different network sizes regardless of the number of cells or UEs, (b) the ability to tackle graph representation tasks, making GNN a powerful tool for extracting useful (often low-dimensional) embedding for the RAN while capturing RAN graph structure (i.e., UE-cell connections), and (c) permutation-invariant processing of graph data (e.g., in aggregating nodes' embeddings), making RAN data processing indifferent to the ordering of cells and UEs. In ~\cite{9679991}, our team developed a GRL-based approach for LB, however, the scope was limited to Best Effort (BE) traffic with no QoS constraints, full-buffer traffic, and an overly simplified network setup.
    
    The main contribution of this paper is a novel QoS-aware LB solution for Open RAN (O-RAN). 
    The solution builds on GRL by modeling RAN states as graphs with heterogeneous nodes (UEs and cells), and treating the LB problem as a Markov Decision Process (MDP). A realistic O-RAN environment is considered  with multi-band cell deployments, non-full buffer mixed traffic including both GBR and BE flows, and support for both intra-band and inter-band handovers. 
    An algorithm is proposed to train the LB agent using a new off-policy dueling Deep Q Network (DQN) that leverages a GNN-based architecture. The proposed algorithm includes a sub-routine to sequentially apply the inference to smaller sub-graphs, rather than processing the entire RAN graph at once, thereby ensuring scalability. 
    The performance of the developed GRL-based solution is compared with two baseline methods: (1) maximum Signal-to-
    Interference-plus-Noise-Ratio (max-SINR), and (2) maximum Reference Signal Received Power (max-RSRP). Simulation results show that the proposed solution yields substantial performance gains, including more than $53\%$ reduction in QoS violations and a fourfold increase in the 5th percentile rate for  BE traffic.\vspace{-.5em}

    \section{System Model}\label{sec:system_model}\vspace{-0em}
    Consider a multi-band O-RAN where each cell site consists of three cells, operating at non-overlapping bands. Let $K$ and $b_k$ denote, respectively, the total number of cells in a set $\mathcal{K}$, and the operating band for cell $k$. Here, our focus is on downlink  with an Orthogonal Frequency-Division Multiple Access
    (OFDMA). At a frequency band $b$, the channel bandwidth is divided equally into $J_b$ subcarriers, each with bandwidth $\omega_b$.  
    
    The O-RAN architecture provides near real-time RAN Intelligent Controller (near-RT RIC) that hosts an LB policy, implemented as an xApp. 
    LB can be achieved by dynamically analyzing the RAN data, and accordingly, moving \emph{cell-edge} UEs from congested cells to nearby under-loaded cells. A UE is located at the cell-edge if the RSRP received from its serving cell is close (e.g., within a $5$ dB difference) to that of the strongest nearby cell. RAN data is sent to the xApp via the E2 interface either periodically or based on event-triggered measurements. UEs connect to one cell at a time and can perform either inter-band or intra-band handovers if instructed by the LB xApp. Let $\mathcal{U}$ be the set of $U$ UEs in the network. The achievable rate at a given transmission time interval (TTI) for UE $u \in \mathcal{U}$ served by cell $k$ will be
    \begin{align}\label{rate}\vspace{-2em}
    	r_{u,k} = \sum_{j=1}^{J_{b_k}}y_{u,k,j} \omega_{b_k}\text{SE}(\gamma_{u,k,j}),
    \end{align}
    where $\gamma_{u,k,j}$ is the SINR, $\text{SE(.)}$ is the spectral efficiency determined based on the rate adaptation mechanism of the cell, and $y_{u,k,j}$ is the resource allocation indicator, i.e., $y_{u,k,j}=1$ if a subcarrier $j$ is allocated to UE $u$ by its serving cell $k$\footnote{\textcolor{black}{Resource allocation is done based on physical resource blocks (PRBs), each comprising a group of subcarriers.}}. 
    \subsection{QoS, Traffic Model, and Scheduling}\label{subsec:traffic_model}
    We consider a generic scenario where majority of UEs have strict GBR requirements while the rest demand BE traffic. Such a mixed-traffic scenario can occur in an enhanced mobile broadband (eMBB) network slice where the network must support GBR transmissions to QoS UEs in presence of background BE traffic. Let $\mathcal{U}^{\text{GBR}}$ and $\mathcal{U}^{\text{BE}}$ denote the subset of UEs demanding GBR and BE traffic, respectively.  To avoid over-complicating the model, each UE has either a GBR or BE (non-GBR) flow, but not both ($\mathcal{U}^{\text{GBR}} \cap \mathcal{U}^{\text{BE}}=\emptyset$).
    
    According to 3GPP, the Guaranteed Flow Bit Rate (GFBR), $r^{\text{GFBR}}$, is referred to as a data rate below which the service is not usable~\cite{3gppTR23501}. In addition, the Maximum Flow Bit Rate (MFBR), $r^{\text{MFBR}} > r^{\text{GFBR}}$, is the highest deliverable rate expected for the QoS flow. That said, QoS for GBR UEs is translated into delivering a data rate that is higher than  $r^{\text{GFBR}}$ and as close as possible to $r^{\text{MFBR}}$. Therefore, we define a \emph{QoS metric} $q_{u,k}$ for GBR UE $u \in \mathcal{U}_{\text{GBR}}$ served by a cell $k$ as
        \begin{align}\label{qos}
    	q_{u,k} =\begin{cases}
    		\min(\bar{r}_{u,k}/r_{u}^{\text{MFBR}}, 1),&\text{if}\,\,\, \bar{r}_{u,k} \geq  r_u^{\text{GFBR}},\\
    		0, &\text{otherwise.}
    	\end{cases}
    \end{align}
    In \eqref{qos}, $\bar{r}_{u,k}$ is an estimate of the ergodic rate calculated by averaging $r_{u,k}$ in \eqref{rate} over an observation window of $T\gg 1$ TTIs. There is no minimum rate constraints for the BE traffic. However, to ensure fairness in  allocating bandwidth resources to GBR and BE traffic, the network must account for the coverage of BE traffic, $c_{\text{BE}}$, defined as
    \begin{align}\label{coverage}
    	c_{\text{BE}} = \inf\left\{\bar{r} | F(\bar{r})\leq 0.05 \right\},
    \end{align}
    where $F(\bar{r})$ represents the cumulative distribution function (CDF) of data rate for  $u \in \mathcal{U}_{\text{BE}}$. \textcolor{black}{The 5th percentile rate, defined as coverage  in~\eqref{coverage}, captures the performance of cell-edge UEs with BE traffic. Since LB is applied to cell-edge UEs, coverage can better capture the LB performance compared to other metrics such as sum throughput.}

    We consider a non-full buffer traffic for each UE $u$ with packet arrival following a Poisson process 
    at the packet data convergence protocol (PDCP) layer. Automatic Repeat Request (ARQ) is considered at the Radio Link Control (RLC) layer. Meanwhile, packets are dropped if the total transmission delay exceeds a delay budget of $\tau_u$ TTIs~\cite{3gppTR23501}. Link adaptation is employed by cells to adjust the rate according to non-ideal periodic channel state information (CSI) feedback received from UEs. Further, resource allocation is performed based on weighted proportional fair (PF) scheduling to account for QoS, channel quality, and fairness. 
    
    \subsection{Problem Formulation}\label{subsec:problem_formulation}
    We formally define an LB policy $\pi(u; \boldsymbol{\theta}): \mathcal{U}^{\text{ce}} \rightarrow \mathcal{K}$, parameterized by $\boldsymbol{\theta}$, as a mapping from the set of cell-edge UEs $ \mathcal{U}^{\text{ce}}\subset \mathcal{U}$ to the set of cells $\mathcal{K}$.  We use $k_u^{\text{sc}}\in \mathcal{K}$ to denote the current serving cell for a UE $u$ before making an LB decision. 
    
    By re-assigning cell edge UEs from an overloaded cell to an appropriate nearby cell, the LB xApp aims  to prevent cell congestion, maximize QoS for the GBR traffic, and improve coverage for BE UEs. To this end, we formulate the following optimization problem:\vspace{-0.5em}
    \begin{subequations}
    \begin{align}
    	&\!\!\!\argmax_{\boldsymbol{\theta}} &&\!\!\!\!\!\!\!\sum_{u \in \mathcal{U}^{\text{GBR}}\cap\mathcal{U}^{\text{ce}}} \!\!\! q_{u,\pi(u;\boldsymbol{\theta})} +\!\!\! \sum_{u \in \mathcal{U}^{\text{GBR}}/\mathcal{U}^{\text{ce}}} \!\!\! q_{u,k_u^{\text{sc}}} +  \alpha c_{\text{BE}}, &&\label{opt_a}\\
    	&\text{s.t.,} &&\!\!\!\!\!\!\!w(\pi(u;\boldsymbol{\theta}))<w(k_u^{\text{sc}}), &&\!\!\!\!\!\!\!\!\!\!\!\!\!\!\!\!\!\!\!\!\forall u \in \mathcal{U}^{\text{ce}}, \label{opt_b}\\
    	&\,&&\!\!\!\!\!\!\!\text{MCS}(u,\pi(u;\boldsymbol{\theta}) )\geq \text{MCS}_{\text{min}}, &&\!\!\!\!\!\!\!\!\!\!\!\!\!\!\!\!\!\!\!\!\forall u \in \mathcal{U}^{\text{ce}},\label{opt_c}\\
    	&\,&&\!\!\!\!\!\!\!\text{RSRP}(u,\pi(u;\boldsymbol{\theta}) )\geq \text{RSRP}_{\text{min}}, &&\!\!\!\!\!\!\!\!\!\!\!\!\!\!\!\!\!\!\!\!\forall u \in \mathcal{U}^{\text{ce}}.\label{opt_d}
    \end{align} 
\end{subequations}
    In \eqref{opt_a}, $\alpha$ is a scaling coefficient to control the tradeoff between maximizing QoS for the GBR traffic versus improving the coverage for BE UEs.  Note that the LB policy impacts all three terms in \eqref{opt_a}. In~\eqref{opt_b}, $w(k)$ represents the bandwidth utilization rate of cell $k$. The  constraint in~\eqref{opt_b} ensures that the LB offloads UEs from a cell with a higher load to a less congested cell. The constraint in~\eqref{opt_c} guarantees that the new serving cell $\pi(u;\boldsymbol{\theta})$ for UE $u$ can support the minimum modulation and coding scheme (MCS), defined by the wideband SINR of the channel. Similarly,  the constraint in~\eqref{opt_d} ensures that the RSRP received from the new serving cell is above the minimum required RSRP. 
    
    Solving the constrained integer programming problem in~\eqref{opt_a}-\eqref{opt_d} is challenging. That is because deriving a closed-form expression for the objective function in~\eqref{opt_a} is not feasible when dealing with the complex realistic network scenario considered in this work. In particular, analytical derivations for the QoS metric $q_{u,k}$ and coverage $c_{\text{BE}}$ are not straightforward in presence of interference, non-full buffer traffic, packet retransmissions, and interdependent performance of UEs in time-varying channels.

    \section{Proposed QoS-Aware Load Balancing based on Graph Reinforcement Learning}\label{proposed}
    
    To solve the proposed problem, we develop a data-driven approach that builds on an off-policy GRL-based learning method while handling QoS constraints in LB, as well as generalization to different network topologies. 
    
    \subsection{RAN graph representation: preliminaries}\label{prelim}
    The RAN environment can be represented as a graph $G=(\mathcal{N}, \mathcal{E}, \boldsymbol{X}_u, \boldsymbol{X}_c)$ with UEs and cells representing its heterogeneous nodes (vertices) in $\mathcal{N}=\mathcal{U}\cup\mathcal{K}$, and access links representing edges in $\mathcal{E}$.  $\boldsymbol{X}_u \in \mathbb{R}^{N \times F_u}$ and $\boldsymbol{X}_c \in \mathbb{R}^{N \times F_c}$ are the input feature matrices where $N=|\mathcal{N}|=U+K$, and $F_u$ and $F_c$ are the number of input features for UE and cell nodes, respectively. We consider two types of edges in $\mathcal{E}$: (a) UE-to-cell edges  in $\mathcal{E}^{\text{uc}}\subset\mathcal{E}$ that represent access links between UEs and cells, (b) cell-to-cell edges in $\mathcal{E}^{\text{cc}}\subset\mathcal{E}$, representing (virtual) links among the cells, helping the LB xApp capture the interdependent performance across the cells when making LB decisions. More concretely, we consider an edge between two cells $k$ and $k'$, if there is at least one cell-edge UE $u \in \mathcal{U}^{\text{ce}}$ for whom the constraint \eqref{opt_d} is satisfied by both cells. Such a cell-to-cell edge naturally indicates a potential load balancing opportunity between cells  $k$ and $k'$. Based on the above definition for $G$, the graph representation for RAN will be dynamic and evolving over time.
    
     \subsection{QoS-aware load balancing as a Markov Decision Process}\label{GRL}
     Using the provided graph representation for the RAN, the LB problem can be translated into a sequential decision making process that modifies graph $G$ over time by changing UE-to-cell edges $\mathcal{E}^{\text{uc}}\subset\mathcal{E}$. In fact, an LB decision by the xApp to offload a cell-edge UE $u$  is equivalent to selecting between two graphs that have identical edges, except the two edges that determine cell association for UE $u$. That said, we can solve the proposed problem in \eqref{opt_a}-\eqref{opt_d} as an MDP $\mathcal{M}=(\mathcal{S}, \mathcal{A}, R, \beta)$, where $\mathcal{S}$ is the state space encompassing the set of all feasible RAN graphs, $\mathcal{A}$ denotes the set of all possible actions that can be selected at various states, $R: \mathcal{S} \times \mathcal{A} \times \mathcal{S} \rightarrow \mathbb{R} $ is the reward function, and $\beta \in \left[0, 1\right]$ is the discount factor.
     
      \noindent\textbf{State:} We define a RAN state $s\triangleq G$ using the graph representation provided in Sec.~\ref{prelim}. At a given LB decision making instance, the sets $\mathcal{E}^{uc}$ and $\mathcal{E}^{cc}$ are determined, respectively, based on the existing UE-cell associations and RSRP measurement reports. Further, the features for UEs and cells are described as follows:
     \begin{itemize}
     	\item \emph{UE features:}  $r_u^{\text{MFBR}}$, $r_u^{\text{GFBR}}$, wideband long-term SINR, UE's average data rate, and the delay budget $\tau_u$.
     	\item \emph{Cell features:} Averaged bandwidth utilization rate of the cell over an observation window of $T$ TTIs, and the average number of  active UEs in the cell.
     \end{itemize}

       \noindent\textbf{Action:} An action is defined as a tuple $a = (u, k', k)\in \mathcal{U}^{\text{ce}} \times \mathcal{K} \times \mathcal{K}$, representing load balancing by re-assigning a UE $u$ from a serving cell $k'$ to a target cell $k$. Each action must be \emph{feasible}, i.e., it should conform to the feasibility constraints in \eqref{opt_b}-\eqref{opt_d} to be qualified for LB. From feasible actions at a given state, one can define $\mathcal{K}^{\text{trg}} \subseteq \mathcal{K}$ and $\mathcal{U}^{\text{trg}}\subseteq \mathcal{U}^{\text{ce}}$, respectively, as  the set of target cells and UEs that can participate in LB. For each feasible LB action $a_i = (u, k', k)$ at a given state graph $G^{(i-1)}$, we can define a feasible \emph{state-action} graph by removing the edge between UE $u$ and its current serving cell $k'$ in $G^{(i-1)}$, while adding a new edge between $u$ and its target cell $k$. Additionally, at each state, we define an auxiliary action as the \emph{stay action}, allowing the agent not to offload any UEs, and thus, stay in the current state. This is motivated by the fact that in some RAN states, forcing LB may not lead to an improved performance in terms of QoS or coverage. 

        \noindent\textbf{Reward:} In order to optimize the LB policy based on the given objective in \eqref{opt_a}, let $f: \mathcal{S}
         \rightarrow \mathbb{R}$ be a mapping as follows:
       \begin{align}\label{reward1}
       	f(G) = \sum_{u \in \mathcal{U}^{\text{GBR}}\cap\mathcal{U}^{\text{ce}}} \!\!\! q_{u,\pi(u;\boldsymbol{\theta})} +\!\!\! \sum_{u \in \mathcal{U}^{\text{GBR}}/\mathcal{U}^{\text{ce}}} \!\!\! q_{u,k_u^{\text{sc}}} +  \alpha \bar{r}_{\text{min, BE}},
       \end{align}
        where $\bar{r}_{\text{min, BE}}$ is the minimum average rate of UEs in $\mathcal{U}^{\text{BE}}$. The first two terms in~\eqref{reward1} aim to increase QoS for the GBR traffic. Additionally, maximizing the third term increases the coverage for the BE traffic. In~\eqref{reward1}, $\bar{r}_{\text{min, BE}}$ is used instead of $c_{\text{BE}}$ since directly estimating the 5th percentile rate from limited measurement samples is not feasible. The reward is then defined as how much $f(.)$ is improved after taking an action. That is, the reward signal at an LB decision step $i$, converting the current state graph $G^{(i-1)}$ to a new graph $G^{(i)}$, is given by:
       \begin{align}\label{reward2}
       	R_i = f\left(G^{(i)}\right) - f\left(G^{(i-1)}\right).
       \end{align}
       The goal of the agent is to find an optimal LB policy $\pi^*(.,\boldsymbol{\theta})$ that maximizes the cumulative discounted return as
     \begin{align}\label{episodic_reward}
     	R_{t:\infty}= \sum_{i=t}^{\infty} \beta^{(i-t)} R_i.
     \end{align} Next, we develop a GRL solution to optimize the LB policy.
    
    \subsection{GNN-based dueling deep Q network} 
    For the LB agent that follows a stochastic policy $\pi$, the state-action and state value functions are defined as follows:
   	\begin{align}
   		Q^{\pi}(s, a)  &= \mathbbm{E}\left[	R_{t:\infty} | s_t=s, a_t=a, \pi \right] ,\\
   		V^{\pi}(s)  &= \mathbbm{E}_{a \sim \pi(s)}\left[Q^{\pi}(s, a) \right].
   	\end{align}
	The state value $V^{\pi}(s)$ determines how good an state $s$ is in terms of accumulating the long-term discounted return, while $Q^{\pi}(s, a)$ measures how good it is to take an action $a$ while being in state $s$. The optimal $Q$ function, $Q^*(s, a) = \max_{\pi} Q^{\pi}(s, a)$ satisfies the Bellman equation:
	\begin{align}\label{bellman}
		Q^*(s, a) =\mathbbm{E}_{s'}\left[R + \beta \max_{a'}Q^*(s',a')|s, a\right],
	\end{align}
	where $R$ is the per step reward in \eqref{reward2}. 
	
	        \begin{figure}
		\centering
		\centerline{\includegraphics[width=8.8cm]{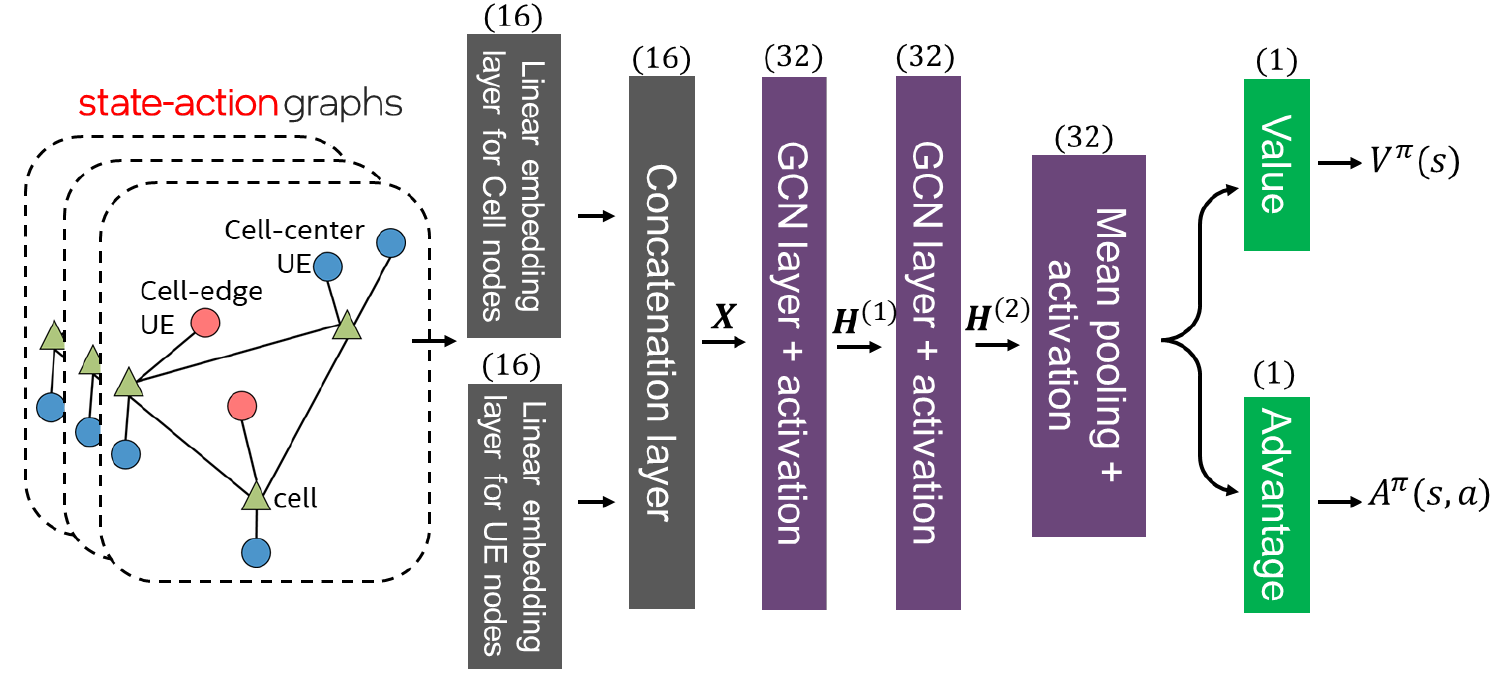}}\vspace{-0em}
		\caption{Proposed GNN-based dueling DQN.}
		\label{model}\vspace{-2em}
	\end{figure}
	
	As motivated earlier, we build on GNN-based deep Q network to estimate the $Q$ values. The proposed Q network architecture is shown in Fig.~\ref{model}. The model includes two initial linear embedding layers to transform the input features of cell nodes ($F_c=2$) and UE nodes ($F_u=5$) into initial embedding vectors of equal length $F_0$, which subsequently are concatenated to construct $\boldsymbol{X} \in \mathbb{R}^{N \times F_0}$. The initial embedding $\boldsymbol{X}$ is then passed through two Graph Convolutional Network (GCN) layers, involving the following layer-wise propagation rule to update the embedding of nodes in $\mathcal{N}$~\cite{kipf2017semisupervised}:
    \begin{align}\label{gcn}
    	\boldsymbol{H}^{(l+1)} = \sigma\left(\tilde{\boldsymbol{D}}^{-1/2}\tilde{\boldsymbol{A}}\tilde{\boldsymbol{D}}^{-1/2}\boldsymbol{H}^{(l)}\boldsymbol{\Theta}^{(l)}\right),
    \end{align}
    where $\boldsymbol{H}^{(l)}\in \mathbb{R}^{N\times F_l}$ is the matrix of activations in layer $l$, with $F_l$ representing the embedding size. Moreover, $\boldsymbol{\Theta}^{(l)}\in \mathbb{R}^{F_l\times F_{l+1}}$ is the learnable parameters of the $l$-th GCN layer, and  $	\boldsymbol{H}^{(l+1)}\in \mathbb{R}^{N\times F_{l+1}}$ is the convolved output embedding matrix (note:  $\boldsymbol{H}^{(0)} = \boldsymbol{X}$). In addition, $\tilde{\boldsymbol{A}} = \boldsymbol{A} + \boldsymbol{I}_N$ with $\boldsymbol{A}\in \mathbb{R}^{N\times N}$ representing the binary adjacency matrix and $ \boldsymbol{I}_N$ denoting the identity matrix. Hence, $\tilde{\boldsymbol{A}}$ represents the adjacency matrix with added self-connections. $\tilde{\boldsymbol{D}}$ is a diagonal degree matrix with elements $\tilde{D}_{ii}=\sum_j \tilde{A}_{ij}$. The GCN layers are followed by a mean pooling layer that performs element-wise averaging over the embeddings of all cell nodes in the graph. The output is then passed into two paths to generate the state value $V^{\pi}(s)$ and the advantage $A^{\pi}(s, a) = Q^{\pi}(s, a) - V^{\pi}(s)$. The state-action values can then be estimated as:
    \begin{align}\label{dueling}
    	Q^{\pi}(s, a) = V^{\pi}(s) +\! \left(\!A^{\pi}(s, a) - \frac{1}{|\mathcal{A}|}\sum_{a'}A^{\pi}(s, a')\right),
    \end{align} 
    where the subtraction in~\eqref{dueling} is to address the issue of identifiability (see~\cite{wang2016duelingnetworkarchitecturesdeep} for details).


    
    \subsection{Proposed GRL-based load balancing algorithm}\label{algo}

     \begin{algorithm}[!t]
    	\footnotesize
    	\caption{Proposed GRL-based load balancing algorithm}\label{Algo:2}
    	\textbf{Inputs:}\,\, RAN graph $G$, DDQN model with parameters $\boldsymbol{\theta}$, set of cell-edge UEs $\mathcal{U}^{\text{cell-edge}}$. \\
    	\textbf{Initialize}\,\, Set the input feature matrices $\boldsymbol{X}_u$ and $\boldsymbol{X}_c$ based on current graph $G$. Identify the set of target cells in $\mathcal{K}^{\text{trg}}$ and UEs in $\mathcal{U}^{\text{trg}}$.
    	\begin{algorithmic}[1]
    		\State Run Algorithm~\ref{Algo:1} to obtain $\mathcal{K}^{\text{sg}}$ and $G^{\text{sg}}$.
    		\State Update features $\boldsymbol{X}^{\text{sg}}$.
    		\If{$\mathcal{K}^{\text{sg}} \neq \emptyset$}
    		\While{episode ends}
    		\State Compute $Q^{\pi}(s, a)$ via~\eqref{dueling}, for all feasible state-action graphs.
    		\State Select action $a^* = \argmax_a Q^{\pi}(s, a)$.
    		\State Perform handover for the UE associated to action $a^*$.
    		\State Observe the performance over $T$ TTIs and calculate reward $R$.
    		\State Update $G$ and store $(s, a, s', R)$ into replay buffer $\mathcal{B}$. 
    		\State Remove offloaded UE from $\mathcal{U}^{\text{trg}}$, and its corresponding target cells \indent $\,\,\,\,$ from $\mathcal{K}^{\text{trg}}$, to prevent frequent handovers.
    		\EndWhile
    		\State Repeat starting from step 1.
    		\Else
    		\State Return.
    		\EndIf
    	\end{algorithmic}
    	\textbf{Output:}\,\,  Updated RAN graph after applying load balancing.
    \end{algorithm}\setlength{\textfloatsep}{0\baselineskip}

Calling on the xApp to perform LB can be done periodically or based on specific events (e.g., handover) in O-RAN. Let $G^{(0)}$ be the initial state of the RAN at the time of calling the LB xApp. The xApp identifies a set of feasible actions and selects one according to its policy $\pi$. After applying the handover based on the selected LB action, the xApp receives the RAN measurements over the next observation window of $T$ TTIs, updates the node features, and selects the next action. Throughout this sequence of LB decisions, hereinafter referred to as an \emph{episode}, the RAN graph evolves as $G^{(0)} \rightarrow G^{(1)} \rightarrow \cdots G^{(m)}$. An episode terminates when there are no more feasible actions to take or if the stay action is selected by the agent (i.e., xApp decides not to move any more cell-edge UEs). As such, the episode length, $m$, varies depending on the network and the agent's policy. The details of this process are summarized in steps $4$-$11$ of  Algorithm~\ref{Algo:2}.

 Applying the proposed framework to real-world O-RAN systems can result in prohibitively large graphs that are difficult to process, thus, hindering the scalability of the proposed solution. To address this challenge, we propose a graph segmentation heuristic in Algorithm~\ref{Algo:1} that allows the LB xApp extract a \emph{sub-graph} $G^{\text{sg}}$ from a larger RAN graph $G$.  
 This sub-routine is then used in step $1$ of Algorithm~\ref{Algo:2}, allowing the proposed method to iteratively  apply LB to the identified sub-graphs with a manageable complexity. 
 
 To train the dueling DQN agent, Algorithm~\ref{Algo:2} is employed by adding $\epsilon$-greedy exploration to the policy. That is, in step $6$, the agent can take a random action with probability $\epsilon$, while following greedy action selection $\argmax_a Q^{\pi}(s, a)$ with probability $1-\epsilon$. The model weights can periodically be updated by minimizing the following loss function at each iteration $i$:
 \begin{align}\label{loss}
 	L(\boldsymbol{\theta}_i) = \mathbb{E}_{(s, a, s', R)\sim \mathcal{B}}\left[(y_{i}^{\text{target}} - Q(s, a; \boldsymbol{\theta}_i))^2\right],
 \end{align}
 with the target value, $y_{i}^{\text{target}}$, be calculated by:
 \begin{align}
 	y_{i}^{\text{target}} = R + \beta \max_{a'}Q(s',a';\boldsymbol{\theta}^{-}),
 \end{align}
 where $\boldsymbol{\theta}^{-}$ denote the parameters of the target network. The model weights  are updated by employing an stochastic gradient descent (SGD) algorithm and applying backpropagation over mini batches of randomly sampled experiences $(s, a, s', R)$ stored in the replay buffer $\mathcal{B}$.

     \begin{algorithm}[!t]
    	\footnotesize
    	\caption{Dynamic RAN subgraph extraction}\label{Algo:1}
    	\textbf{Inputs:}\,\, RAN graph $G$, set of target cells in $\mathcal{K}^{\text{trg}}$ and UEs in $\mathcal{U}^{\text{trg}}$.\\
    	\textbf{Initialize}\,\, List of selected cells to be included in the subgraph $ \mathcal{K}^{\text{sg}} =\emptyset$. 
    	\begin{algorithmic}[1]
    		\For{each cell-edge UE $u \in \mathcal{U}^{\text{trg}}$}
    		\State Find the set of target cells $\mathcal{K}_u^{\text{trg}}\subseteq \mathcal{K}^{\text{trg}}$.
    		\EndFor
    		\For{target cell $k \in \mathcal{K}^{\text{trg}}$}
    		\State Initialize a set of tentative cells in the subgraph as  $\mathcal{K}_{\text{tn}}^{\text{sg}}=\emptyset$.
    		\For{each cell-edge UE $u \in \mathcal{U}^{\text{trg}}$}
    		\State If $k \in \mathcal{K}_u^{\text{trg}}$, then set $\mathcal{K}_{\text{tn}}^{\text{sg}} \leftarrow \mathcal{K}_{\text{tn}}^{\text{sg}} \cup \mathcal{K}_u^{\text{trg}}$.
    		\EndFor
    		\If{$|\mathcal{K}_{\text{tn}}^{\text{sg}}| > |\mathcal{K}^{\text{sg}}|$}
    		\State $\mathcal{K}^{\text{sg}} \leftarrow \mathcal{K}_{\text{tn}}^{\text{sg}}$.
    		\EndIf
    		\EndFor
    	\end{algorithmic}
    	\textbf{Output:}\,\,  $\mathcal{K}^{\text{sg}}$ and its corresponding subgraph $G^{\text{sg}}$ constructed from cells in $\mathcal{K}^{\text{sg}}$ and their associated UEs.
    \end{algorithm}\setlength{\textfloatsep}{0\baselineskip}

    \section{Simulation Results}\label{sec:sim_results}
    Detailed simulations are conducted using a Python system-level simulator which supports 3GPP compliant channel models, non-ideal CSI feedback, non-full buffer traffic, link adaptation, RLC re-transmissions, and weighted PF scheduling.  The network consists of seven hexagonal cell sites located in a 1.5km $\times$ 1.5km simulation area, with a minimum inter-site distance of $500$m. Each site includes three omni cells operating at distinct frequency bands reported in Table~\ref{tab1}. For each deployment, UEs are distributed  randomly within the simulation area. Key parameters related to the wireless network are selected based on 3GPP~\cite{3gppTR23501} and listed in Table~\ref{tab1}.
    
     \begin{table}
    	\footnotesize
    	\centering
    	\caption{Simulation Parameters}\vspace{-0em}
    	\begin{tabular}{|d{.40}|d{.45}|}
    		\hline
    		\textbf{Parameter}& \textbf{Value} \\
    		\hline
    		
    		Number of UEs ($U$)& $\left\{35, 42, \cdots, 70\right\}$  \\
    		\hline
    		
    		Number of cells ($K$)& $21$  \\
    		\hline
    		
    		Operating bands (carrier frequency, channel bandwidth)& \begin{itemize}
    			\item n29 ($725$ MHz, $10$ MHz)
    			\item  n66 ($2190$ MHz,  $5$ MHz) 
    			\item  n1 ($2140$ MHz,  $5$ MHz)\vspace{-1em} 
    		\end{itemize}  \\
    		\hline
    		
    		Subcarrier spacing& $15$ kHz  \\
    		\hline
    		
    		TTI duration& $1$ ms \\
    		\hline
    		
    		Cell transmit power& $44$ dBm  \\
    		\hline
    		
    		QoS flow 5QI& for GBR: [2, 3, 67], for BE: 9  \\
    		\hline
    		
    		Input data rate per UE& \begin{itemize}
    			\item GBR: $\!\!\left\{0.8, 1, \cdots, 16 \right\}$ Mbps
    			\item BE: $\!\!\left\{2, 2.4, \cdots, 8 \right\}$ Mbps\vspace{-0.6em}
    		\end{itemize}  \\
    		\hline
    		
    		
    		Delay threshold ($\tau_u$)& Based on 5QI table in~\cite{3gppTR23501}  \\
    		
    		\hline
    		$\text{MCS}_{\text{min}}$  index, $\text{RSRP}_{\text{min}}$& $0$,  $-120$ dBm \\

    		\hline
    	\end{tabular}
    	\label{tab1}\vspace{-1.8em}
    \end{table}
    
    The proposed GRL method is implemented in PyTorch using the PyTorch Geometric (PyG) library. The numbers above each layer in Fig.~\ref{model} show the output length of embedding vector for each node. Leaky Rectified Linear Unit (ReLU) is used as the activation function. The model is trained over deployments with random selection of the number of UEs, UEs' locations, and traffic parameters. The MFBR $r^{\text{MFBR}}$ is considered equal to the UE's incoming rate, randomly selected from values reported  for each traffic class in Table~\ref{tab1}. Moreover, the GFBR is set as $r^{\text{GFBR}} = \rho r^{\text{MFBR}}$ where $\rho$ takes values $0.6$, $0.9$, and $0.8$, respectively, for QoS flows with 5QIs $2$, $3$, and $67$. QoS and BE flows are assigned based on a $3:1$ ratio. Adam optimizer is used along with a learning rate scheduler, using $0.01$ as initial learning rate.  Each training step involves one round of model update based on SGD over a mini-batch of $64$ experiences randomly sampled from the replay buffer. Moreover, $\beta=0.999$ and the target network is updated  after every $10$ episodes. We compare the proposed GRL-based method with two widely-used baseline approaches, namely, Max-RSRP and Max-SINR.

    
    \begin{figure}
    	\centering
    	\centerline{\includegraphics[width=8cm]{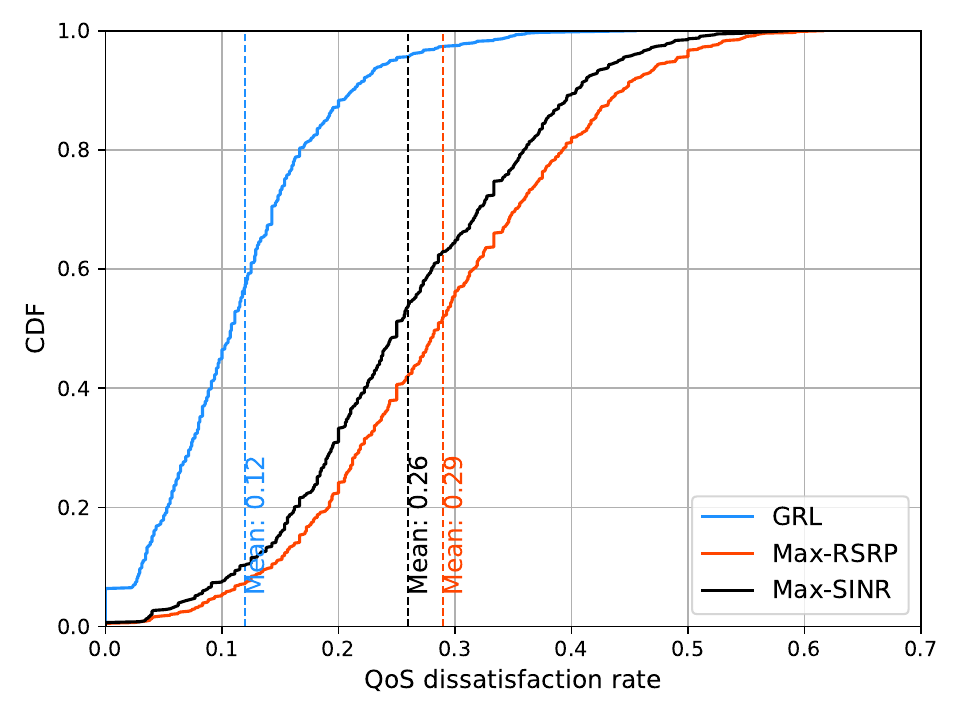}}\vspace{-0.7em}
    	\caption{CDF of the QoS dissatisfaction rate.}
    	\label{fig1}\vspace{-0em}
    \end{figure}
    
     \begin{figure}[!]\vspace{-1em}
    	\centering
    	\centerline{\includegraphics[width=8cm]{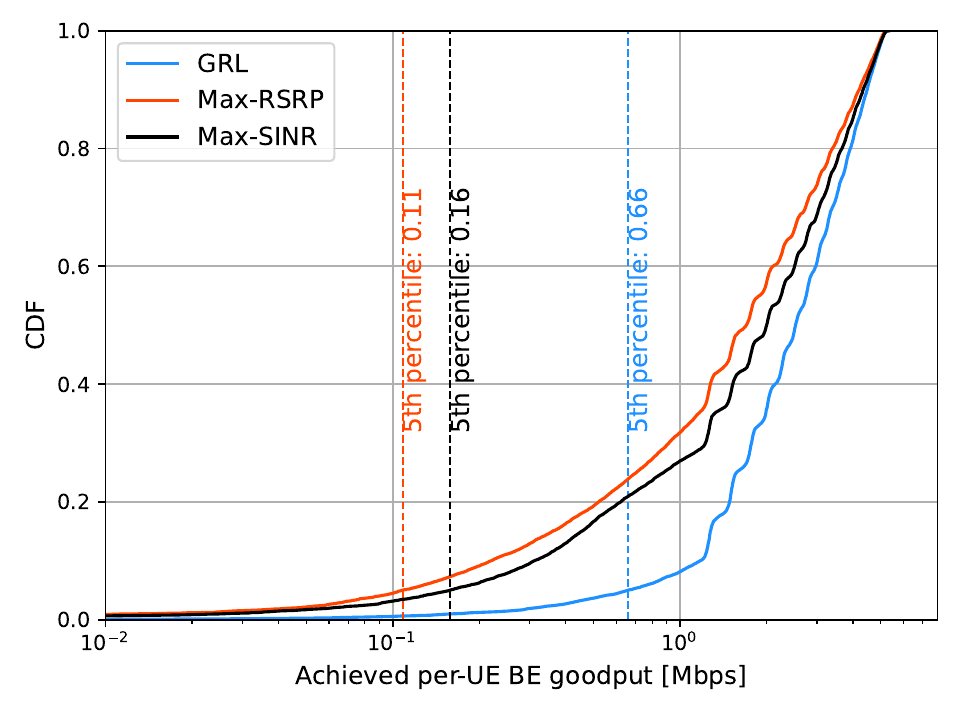}}\vspace{-0.6em}
    	\caption{CDF of the  achieved goodput per BE UE in Mbps.}
    	\label{fig2}\vspace{-0.5em}
    \end{figure}
    
\begin{figure}\vspace{-1em}
    	\centering
    	\centerline{\includegraphics[width=8cm]{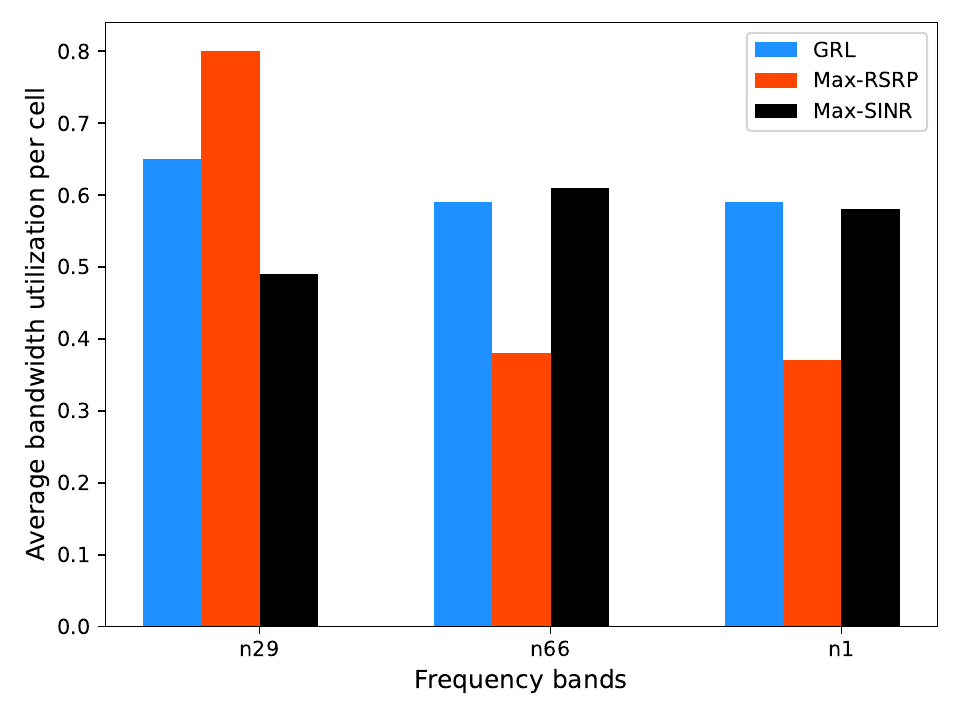}}\vspace{-0.8em}
    	\caption{Comparison of average bandwidth utilization rate per cell.}
    	\label{fig3}\vspace{-0.1em}
    \end{figure}

    Figure~\ref{fig1} compares the CDF of  QoS dissatisfaction rate for the three methods. The QoS dissatisfaction rate is calculated as the number of GBR UEs not meeting their GFBR constraint divided by the total number of GBR UEs. Fig.~\ref{fig1}  shows that the GRL-based method can minimize the average QoS dissatisfaction rate by $53\%$ compared to the baselines. The reason that the average QoS dissatisfaction rate is not zero is because, as shown in Fig.~\ref{fig3}, the  results are obtained for high-traffic network scenarios with an average (post LB) bandwidth utilization of around $60\%$ per cell.

    Figure~\ref{fig2} shows the performance improvements for the BE traffic, yielded by the GRL approach. The figure compares the CDF of per-UE achieved goodput across BE UEs. The vertical lines show the coverage metric, $c_{\text{BE}}$ defined in \eqref{coverage}. Fig.~\ref{fig2} shows that the proposed method can improve the coverage for BE UEs by more than $4$x compared to the baselines.
    
    Figure~\ref{fig3} compares the three methods, in terms of their LB performance, by evaluating the average bandwidth utilization rate per cell. The max-RSRP method assigns most of the traffic to cells at n29 band, due to a more favorable path loss compared to other bands. Such an unbalanced load allocation by max-RSRP method results in cell congestion at n29. Meanwhile, the max-SINR method achieves a more uniform load allocation across the bands by assigning more UEs to n66 and n1 cells (due to less interference and higher average SINR), leading to outperforming the max-RSRP method. However, when compared to the proposed approach, max-SINR method under-utilizes the n29 band.
    Figs.~\ref{fig1}-\ref{fig3} show that the proposed GRL-based LB method can properly distribute the traffic across all three bands while outperforming the baseline methods in terms of QoS satisfaction and service coverage. 
    
    
    


    \section{Conclusions}\label{sec:conclusions}
    In this paper, we have developed a new QoS-aware load balancing method in multi-band O-RAN, while considering both GBR and BE traffic. We have formulated an optimization problem to jointly maximize QoS  and coverage, respectively, for the GBR and BE traffic. We have solved the problem by proposing a new GRL-based algorithm that trains the LB xApp as a dueling DQN agent where the Q network constitutes a GNN. We have conducted thorough evaluations to assess the performance based on several metrics including QoS dissatisfaction rate, coverage, and load balancing. Simulation results have shown the proposed GRL-based solution completely outperforms the baseline methods across all performance metrics.

    \bibliographystyle{IEEEtran}
    \bibliography{bib/references}

\begin{thebibliography}{10}
\providecommand{\url}[1]{#1}
\csname url@samestyle\endcsname
\providecommand{\newblock}{\relax}
\providecommand{\bibinfo}[2]{#2}
\providecommand{\BIBentrySTDinterwordspacing}{\spaceskip=0pt\relax}
\providecommand{\BIBentryALTinterwordstretchfactor}{4}
\providecommand{\BIBentryALTinterwordspacing}{\spaceskip=\fontdimen2\font plus
\BIBentryALTinterwordstretchfactor\fontdimen3\font minus
  \fontdimen4\font\relax}
\providecommand{\BIBforeignlanguage}[2]{{%
\expandafter\ifx\csname l@#1\endcsname\relax
\typeout{** WARNING: IEEEtran.bst: No hyphenation pattern has been}%
\typeout{** loaded for the language `#1'. Using the pattern for}%
\typeout{** the default language instead.}%
\else
\language=\csname l@#1\endcsname
\fi
#2}}
\providecommand{\BIBdecl}{\relax}
\BIBdecl

\bibitem{8869705}
W.~Saad, M.~Bennis, and M.~Chen, ``A vision of {6G} wireless systems:
  Applications, trends, technologies, and open research problems,'' \emph{IEEE
  Network}, vol.~34, no.~3, pp. 134--142, 2020.

\bibitem{8255801}
M.~M. Hasan, S.~Kwon, and J.-H. Na, ``{Adaptive Mobility Load Balancing
  Algorithm for LTE Small-Cell Networks},'' \emph{IEEE Transactions on Wireless
  Communications}, vol.~17, no.~4, pp. 2205--2217, 2018.

\bibitem{10659897}
Y.-J. Cho, H.-M. Yoo, K.-S. Kim, J.~Na, and E.-K. Hong, ``{Practical Load
  Balancing Algorithm for 5G Small Cell Networks Based on Real-World 5G Traffic
  and O-RAN Architecture},'' \emph{IEEE Access}, vol.~12, pp.
  121\,947--121\,957, 2024.

\bibitem{DEMIRCI2018140}
İlhan Demirci and Ömer Korçak, ``{Cell breathing algorithms for load
  balancing in Wi-Fi/cellular heterogeneous networks},'' \emph{Computer
  Networks}, vol. 134, pp. 140--151, 2018.

\bibitem{10723743}
A.~Manan, S.~M. Shahid, S.~Kim, and S.~Kwon, ``{Load Balancing With Traffic
  Splitting for QoS Enhancement in 5G HetNets},'' \emph{IEEE Transactions on
  Network Science and Engineering}, vol.~11, no.~6, pp. 6272--6284, 2024.

\bibitem{10615828}
S.~A. Lahham, D.~Wu, E.~Hossain, X.~Liu, and G.~Dudek, ``{Probabilistic
  Mobility Load Balancing for Multi-Band 5G and Beyond Networks},'' in
  \emph{2024 IEEE ICC Workshops}, 2024, pp. 1673--1678.

\bibitem{9740126}
E.~Gures, I.~Shayea, M.~Ergen, M.~H. Azmi, and A.~A. El-Saleh, ``Machine
  learning-based load balancing algorithms in future heterogeneous networks: A
  survey,'' \emph{IEEE Access}, vol.~10, pp. 37\,689--37\,717, 2022.

\bibitem{10463334}
M.~Shabbir, S.~Kandeepan, A.~Al-Hourani, and W.~Rowe, ``Lstm based proactive
  access point selection and mobility load balancing for ultra-dense
  networks,'' in \emph{2024 International Conference on Artificial Intelligence
  in Information and Communication (ICAIIC)}, 2024, pp. 452--458.

\bibitem{8796404}
Y.~Xu, W.~Xu, Z.~Wang, J.~Lin, and S.~Cui, ``Load balancing for ultradense
  networks: A deep reinforcement learning-based approach,'' \emph{IEEE Internet
  of Things Journal}, vol.~6, no.~6, pp. 9399--9412, 2019.

\bibitem{9369525}
G.~Alsuhli, H.~A. Ismail, K.~Alansary, M.~Rumman, M.~Mohamed, and K.~G. Seddik,
  ``Deep reinforcement learning-based cio and energy control for lte mobility
  load balancing,'' in \emph{2021 IEEE 18th Annual Consumer Communications \&
  Networking Conference}, 2021, pp. 1--6.

\bibitem{9637122}
P.~E. Iturria-Rivera and M.~Erol-Kantarci, ``Qos-aware load balancing in
  wireless networks using clipped double q-learning,'' in \emph{IEEE 18th
  International Conference on Mobile Ad Hoc and Smart Systems}, 2021.

\bibitem{9679991}
O.~Orhan, V.~N. Swamy, T.~Tetzlaff, M.~Nassar, H.~Nikopour, and S.~Talwar,
  ``{Connection Management xAPP for O-RAN RIC: A Graph Neural Network and
  Reinforcement Learning Approach},'' in \emph{20th IEEE International
  Conference on Machine Learning and Applications}, 2021.

\bibitem{lin2023}
\BIBentryALTinterwordspacing
X.~Lin, ``{Artificial Intelligence in 3GPP 5G-Advanced: A Survey},'' 2023.
  [Online]. Available: \url{https://arxiv.org/abs/2305.05092}
\BIBentrySTDinterwordspacing

\bibitem{zhao2023}
\BIBentryALTinterwordspacing
J.~Zhao and C.~Yang, ``Graph reinforcement learning for radio resource
  allocation,'' 2023. [Online]. Available:
  \url{https://arxiv.org/abs/2203.03906}
\BIBentrySTDinterwordspacing

\bibitem{3gppTR23501}
3GPP, ``{System architecture for the 5G System (5GS)},'' TS 23.50, 2020.

\bibitem{kipf2017semisupervised}
T.~N. Kipf and M.~Welling, ``{Semi-Supervised Classification with Graph
  Convolutional Networks},'' in \emph{International Conference on Learning
  Representations}, 2017.

\bibitem{wang2016duelingnetworkarchitecturesdeep}
\BIBentryALTinterwordspacing
Z.~Wang, T.~Schaul, M.~Hessel, H.~van Hasselt, M.~Lanctot, and N.~de~Freitas,
  ``{Dueling Network Architectures for Deep Reinforcement Learning},'' 2016.
  [Online]. Available: \url{https://arxiv.org/abs/1511.06581}
\BIBentrySTDinterwordspacing

\end{thebibliography}

\end{document}